\documentclass[10pt,twocolumn,letterpaper]{article}

\usepackage[pagenumbers]{cvpr} 

\definecolor{cvprblue}{rgb}{0.21,0.49,0.74}
\usepackage[pagebackref,breaklinks,colorlinks,allcolors=cvprblue]{hyperref}

\usepackage{booktabs}
\usepackage{nicematrix}
\usepackage{algorithm}
\usepackage{algorithmic}
\usepackage[frozencache,cachedir=.]{minted}
\usepackage{multirow}
\usepackage{setspace}
\usepackage{bm}
\usepackage{amsmath}
\usepackage{nccmath}
\usepackage{adjustbox}

\newcommand\ours{RID}
\def\paperID{2480} 
\def\confName{CVPR}
\def\confYear{2025}

\title{Reward Incremental Learning in Text-to-Image Generation}

\author{Maorong Wang$^{1,2}$ \quad Jiafeng Mao$^{1}$ \quad Xueting Wang$^{1\footnotemark[1]}$ \quad Toshihiko Yamasaki$^{2}$ \vspace{0.3em} \\
{\normalsize $^1$CyberAgent} \quad
{\normalsize $^2$The University of Tokyo} \quad \\
{\normalsize {\{ma\_wang, yamasaki\}@cvm.t.u-tokyo.ac.jp}, {\{jiafeng\_mao, wang\_xueting\}@cyberagent.co.jp}}
}

\begin{document}

\maketitle

\renewcommand{\thefootnote}{\fnsymbol{footnote}}
\footnotetext[1]{Corresponding Author}
\renewcommand{\thefootnote}{\arabic{footnote}}
\footnotetext[1]{This work was conducted while the first author was doing internship at CyberAgent.}

\begin{abstract}
The recent success of denoising diffusion models has significantly advanced text-to-image generation. While these large-scale pretrained models show excellent performance in general image synthesis, downstream objectives often require fine-tuning to meet specific criteria such as aesthetics or human preference. Reward gradient-based strategies are promising in this context, yet existing methods are limited to single-reward tasks, restricting their applicability in real-world scenarios that demand adapting to multiple objectives introduced incrementally over time. In this paper, we first define this more realistic and unexplored problem, termed \textbf{R}eward \textbf{I}ncremental \textbf{L}earning (RIL), where models are desired to adapt to multiple downstream objectives incrementally. Additionally, while the models adapt to the ever-emerging new objectives, we observe a unique form of catastrophic forgetting in diffusion model fine-tuning, affecting both metric-wise and visual structure-wise image quality. To address this catastrophic forgetting challenge, we propose \textbf{R}eward \textbf{I}ncremental \textbf{D}istillation~(\ours), a method that mitigates forgetting with minimal computational overhead, enabling stable performance across sequential reward tasks. The experimental results demonstrate the efficacy of~\ours~in achieving consistent, high-quality generation in RIL scenarios. The source code of our work will be publicly available upon acceptance. 
\end{abstract}
\vspace{-14pt}


\section{Introduction}
\label{sec:intro}

Recently, we have witnessed the remarkable success of content generation, driven by denoising diffusion models~\cite{ddpm, ldm, ramesh2021zero, ding2021cogview, zhou2022towards, ramesh2022hierarchical}. 
While the large-scale pretrained diffusion models yield excellent performance in synthesizing general high-quality images, researchers have proposed various fine-tuning strategies to adapt these models for specific downstream objectives (\textit{e.g.}, aesthetic, compressibility, human preference, etc.). Existing fine-tuning strategies can be classified into four categories: prompt engineering~\cite{prompt_engineering}, supervised fine-tuning~\cite{lee2023aligning}, reinforcement learning~\cite{black2023training}, and reward gradient-based methods~\cite{imagereward, alignprop, draft}. Among them, reward gradient-based strategies have achieved notable success due to their strong performance and computational efficiency. In this paper, we focus on fine-tuning the text-to-image diffusion models with the reward gradient-based method.



While the existing reward gradient-based methods have shown effectiveness, they generally focus on optimizing only a single reward task. However, in practical applications, generative models must adapt to multiple objectives introduced incrementally over time. For instance, as in Fig.~\ref{fig:RILexample}, a model initially fine-tuned to enhance aesthetic quality may later need to align with human preferences and incorporate compressibility constraints. These incremental adaptations are essential to ensure models can evolve alongside changing requirements. However, we observed that the current methods, when directly applied to sequential task adaptations, suffer from significant performance degradation, limiting their practical applicability. 

Inspired by incremental learning approaches from the continual learning community, we conceptualize this challenge as the \textit{Reward Incremental Learning} (RIL) problem. To the best of our knowledge, this problem has not been formally identified or explored in prior research, despite its relevance to real-world applications. In this paper, we aim to address the RIL problem by proposing a method to incrementally fine-tune text-to-image diffusion models across multiple objectives, balancing adaptability to new tasks with robustness to previously learned ones.

While addressing the RIL problem, we identify a catastrophic forgetting phenomenon in diffusion model fine-tuning, analogous to the widely studied problem in classification tasks~\cite{goodfellow2013empirical, mccloskey1989catastrophic}. Unlike traditional forgetting, which is marked by a drop in prediction accuracy, forgetting in diffusion fine-tuning occurs both metric-wise and visual structure-wise. Metric-wise forgetting (cf. Table~\ref{tab:forgetting}) degrades overall generation quality, including naturalness and text-to-image alignment, as reflected by metrics such as zero-shot MS-COCO FID~\cite{heusel2017gans} and CLIP score~\cite{clipscore}, respectively. Visual structure-wise forgetting disrupts image composition and leads to alterations in objective irrelevant details like the background and the visual form of the object. We spot such a forgetting issue is severe in the adapted baseline method~\ref{subsec:baseline}, and becomes more pronounced as additional tuning tasks are introduced, as shown in Fig.~\ref{fig:RILexample}.




In this work, we first define the RIL setting and quantify the catastrophic forgetting problem in diffusion-based generation scenarios. Then, we explore the performance and limitations of simply adapting the current state-of-the-art methods~\cite{alignprop, draft} to the RIL problem with the proposed LoRA adapter group (denoted as the baseline method in Fig.~\ref{fig:RILexample}). Moreover, we propose \textbf{R}eward \textbf{I}ncremental \textbf{D}istillation~(\ours), a method to more effectively tackle the RIL problem by taking advantage of the proposed LoRA adapter group and momentum distillation techniques to alleviate the catastrophic forgetting issue in the fine-tuning process.~\ours~can achieve high-quality generation results in the RIL setting with negligibly small extra computational overhead (only 2\% extra diffusion steps). 

Our main contribution can be summarized as follows: 

\begin{itemize}
    \item We define the practical but unexplored RIL problem, where the diffusion models are required to adapt incrementally to multiple downstream objectives. 
    
    \item We formally identify and quantify the catastrophic forgetting issue in fine-tuning text-to-image diffusion models, showing its impact on generation quality at both metric and visual structure levels.

    \item We propose~\ours, an approach designed to mitigate catastrophic forgetting with minimal computational overhead. It achieved more consistent and satisfactory generation in both metrics and visual performance across extensive RIL task sequences.
    
\end{itemize}

\begin{table}[]
\begin{center}
    \resizebox{\linewidth}{!}{
    \begin{tabular}{cccc}
\toprule
\multirow{2}{*}{Method} & Target Metric & \multicolumn{2}{c}{General Metrics} \\
\cmidrule(lr){2-2}
\cmidrule(lr){3-4}
 & Aesthetic score $\uparrow$ & CLIP score $\uparrow$ & Zero-shot FID $\downarrow$ \\
\midrule
SD V1.5            & 5.23 & 25.77 & 71.54\\
\midrule
Baseline           & 6.07(+0.84) & 22.74(-3.03) & 85.53(+13.99)\\
\ours              & 6.03(+0.80) & \textbf{25.76(-0.01)} & \textbf{79.40(+7.86)}\\
\bottomrule
\end{tabular}

    }
    \vspace*{-5pt}
    \caption{Metric-wise forgetting in the fine-tuning pretrained diffusion model with aesthetic rewards. The numbers in the parentheses indicate the performance difference compared with the original Stable Diffusion V1.5. To establish the baseline method, we adapted the current state-of-the-art methods~\cite{draft, alignprop} in diffusion fine-tuning to the RIL setting with the proposed LoRA adapter group (cf. Sec.~\ref{subsec:baseline}). Although both methods achieve a similar targeted objective score (Aesthetic score), the baseline method suffers from a huge deterioration in both the CLIP score and Zero-shot COCO FID, indicating a drop in overall generation quality, while the forgetting of RID is greatly alleviated.}
    \vspace{-22pt}
    \label{tab:forgetting} 
\end{center}
\end{table}

\vspace{-4pt}
\section{Related Work}
\label{sec:related}

\noindent\textbf{Diffusion fine-tuning.}
Text-to-image synthesis has achieved success in generating realistic, high-fidelity images, driven by denoising diffusion models~\cite{ddpm, ldm}. 
However, in real-world applications, it is necessary to fine-tune these pretrained models to align with various downstream objectives, such as aesthetics, text-image alignment, or user preferences. Current fine-tuning strategies fall into four categories: prompt engineering~\cite{prompt_engineering}, supervised fine-tuning~\cite{lee2023aligning}, reinforcement learning~\cite{black2023training}, and reward gradient-based methods~\cite{imagereward, alignprop, draft}.

Prompt engineering-based methods~\cite{prompt_engineering} solve the problem by designing and crafting descriptive prompts to enhance the generation quality on given downstream objectives. Since these strategies do not require fine-tuning the diffusion model, it is more computationally efficient but yields limited performance. Supervised fine-tuning strategies~\cite{lee2023aligning} tune the pretrained diffusion models on human-curated datasets (\textit{e.g.}, LAION Aesthetics) to improve the generation quality on the desired objective. However, collecting such datasets can be laborious. Reinforcement learning-based strategies~\cite{black2023training} conquer the necessity of the training dataset needed in the fine-tuning process. Such methods use different reward sources (\textit{i.e.,} human rating, reward model, etc.) to generate scaler reward values and use policy gradients in the fine-tuning process. However, reinforcement learning suffers from high computation cost and limited performance, caused by the high variance of the gradients. Reward gradient-based methods~\cite{imagereward, alignprop, draft} propose to use differentiable reward functions and their gradient to fine-tune the generative model, and it gained attention for its performance and computation efficiency. 


\noindent\textbf{Continual learning and catastrophic forgetting.}
Continual Learning~\cite{wang2024comprehensive, chen2022lifelong, de2021continual, mai2022online} tackles the problem of learning from a sequence of tasks, where the data distribution and learning objective change over time. The core challenge in continual learning is the catastrophic forgetting~\cite{goodfellow2013empirical, mccloskey1989catastrophic} problem, which inevitably happens when the model tries to adapt to new learning objectives. Most existing continual learning research focuses on image classification tasks, where catastrophic forgetting manifests as a decline in accuracy when evaluated with previous test datasets. 

In contrast to the conventional concept of catastrophic forgetting, the RIL setting introduces a unique manifestation of catastrophic forgetting. In diffusion-based text-to-image generation, forgetting extends beyond a mere drop in overall accuracy and is observed as a performance deterioration in the generation results. As mentioned in Sec.~\ref{sec:intro}, such deterioration occurs both in metrics and visual structure. Our work addresses this distinct form of catastrophic forgetting by exploring momentum distillation-based strategies designed to preserve image quality and task consistency across incremental tuning stages.

\section{Reward Incremental Learning}
\label{sec:preliminary}

\begin{figure*}[t]
  \centering
   \includegraphics[width=.95\linewidth]{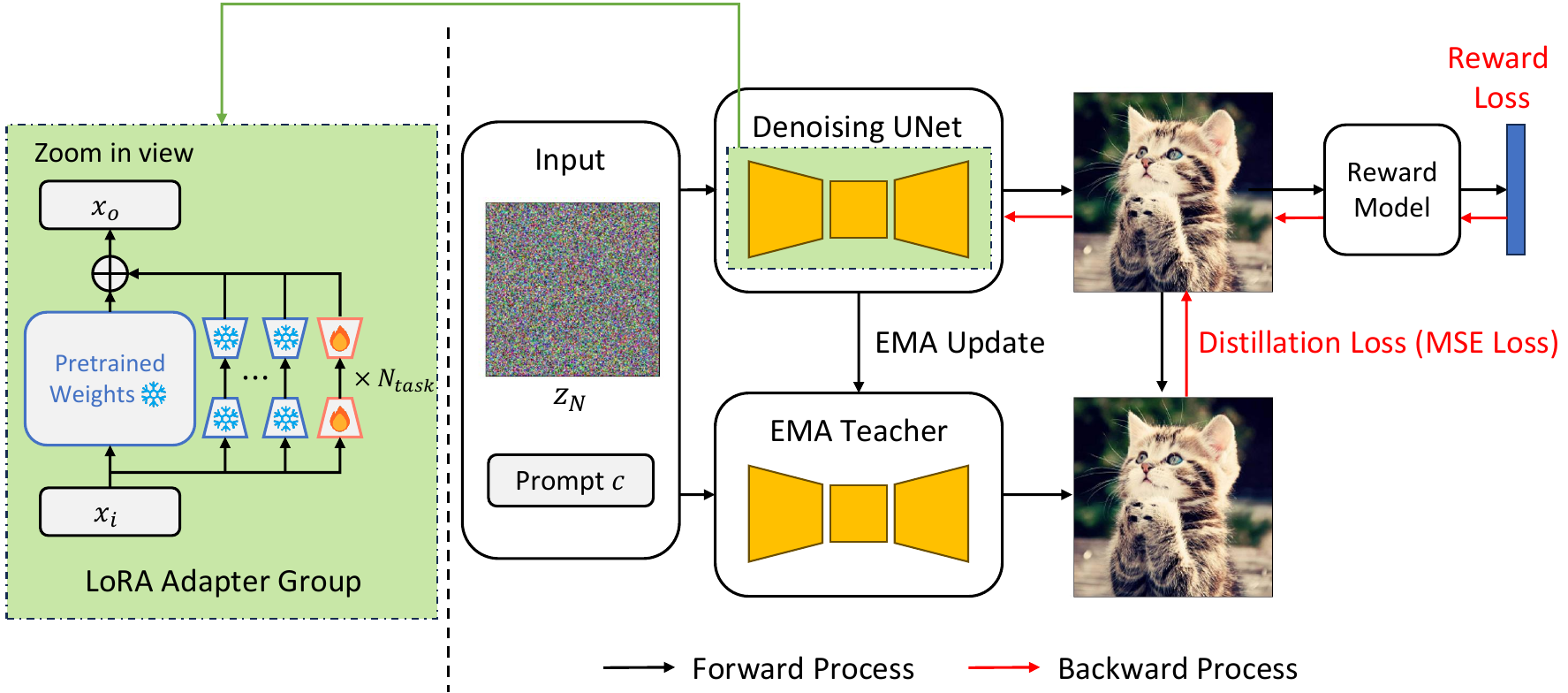}
   \vspace*{-10pt}
   \caption{The overview of our proposed~\ours.~\ours~has two main components: LoRA adapter group and momentum distillation. Through the combined use of group LoRA adapter and EMA distillation,~\ours~achieves improved robustness against forgetting, generating images that not only align well with target objectives but also maintain high general quality.}
   \label{fig:teaser}
    \vspace{-14pt}
\end{figure*}

\subsection{Problem Formulation}
\label{subsec:ril}
Different from the conventional reward fine-tuning methods that only focus on a single reward objective, we propose a more realistic and challenging setting called \textit{Reward Incremental Learning} (RIL).
The main goal of the RIL task is to fine-tune the diffusion model across a sequence of reward tasks incrementally, achieving adaptability to new objectives while preserving previously learned knowledge. For each reward task $t \in \{1,...,T\}$, the objective of the current task is to optimize the image $\bm{z}_0$ generated by the diffusion model over $N$ denoising steps, evaluated with the reward function $R_t(\cdot)$: 
\begin{equation}
\begin{aligned}
\max_{\theta}~~\sum_{c \in C_{train}} R_t(f_\theta(\bm{z}_N | c)), 
\end{aligned}
\label{eq:ril}
\end{equation}
where $\theta$ are the parameters in the diffusion model $f_\theta$ which is conditioned by the text prompt $c$ in the training dataset $C_{train}$, and $\bm{z}_N \sim \mathcal{N}(0,\mathbf{I})$ is the initial noise sampled from the Gaussian distribution. 

\noindent\textbf{Differentiable Reward Functions.}
We introduce the differentiable reward function $R_t(\cdot)$ required for the gradient-based fine-tuning process. In this paper, we evaluate three key reward tasks: aesthetic quality, human preference, and compressibility, which serve as foundational objectives in existing gradient-based reward fine-tuning methods.

\begin{itemize}
    \item \textbf{Aesthetic Reward.} Following~\cite{alignprop, draft}, we use the LAION Aesthetic predictor v2~\cite{laionaesthetics2022} to generate the aesthetic reward. The predictor is trained on the combination of SAC~\cite{SAC}, LAION-Logos~\cite{laionaesthetics2022} and AVA~\cite{AVA} dataset, which contains a total of 441,000 human-rated images scored from 1 to 10, where a score of 10 indicates the best aesthetic quality. Since the aesthetic predictor trains an MLP based on the CLIP~\cite{clip} features, this reward function is differentiable. 

    \item \textbf{Human Preference Reward.} Similar to the aesthetic reward, we use Human Preference Score v2 (HPSv2) predictor~\cite{hps} to generate the rewards for the fine-tuning process. HPSv2 is trained on the Human Preference Dataset (HPD) that comprises 798,090 human preference choices on 433,760 generated image pairs. Unlike the LAION aesthetic predictor, the HPSv2 predictor also takes the prompts used in image generation into its scoring process. Since the HPSv2 model is fine-tuned from the CLIP model on the Human Preference Dataset (HPD), it is differentiable. 

    \item \textbf{Compressibility Reward.} Because the file size is not a differentiable metric for use as the reward function in fine-tuning, we follow~\cite{draft} in using the following reward function: 
    \begin{equation}
    \begin{aligned}
    R(img) = - ||img - C(img)||^2,
    \end{aligned}
    \label{eq:compress}
    \end{equation}
    where $C(\cdot)$ is the compression algorithms implemented with the differentiable JPEG approximation~\cite{diffjpeg}. This reward function encourages the diffusion model to generate simpler images that resemble its compressed counterpart. 
\end{itemize}   

\noindent\textbf{Adaptability-robustness tradeoff.} 
Similar to other incremental learning tasks, the core challenge of the RIL problem is achieving an effective adaptability-robustness tradeoff. This requires the diffusion model to learn new knowledge (novel target objectives) while retaining previously acquired knowledge, including overall generation quality and past objectives.

\subsection{Quantifying the Forgetting}
\label{subsec:forgetting}
The forgetting measure (FM)~\cite{chaudhry2018riemannian} used in conventional continual learning settings (image classification tasks)~\cite{hsu2018re, van2019three}, such as class-incremental learning,
is defined as the performance deterioration in accuracy on historical datasets. Unlike conventional FM, we measure the forgetting in the RIL problem as the task-wise deterioration when tested with the objective evaluation metrics as in Sec.~\ref{subsec:setup}. Apart from the target objectives, we also evaluate the forgetting of the CLIP score and zero-shot MS-COCO FID to measure the forgetting of the general quality of the generated image. 

Formally, for a given task $t \in \{1,...,T\}$, to maintain consistency in presentation where lower forgetting values indicate better performance, we calculate the forgetting $F_t$ depending on whether higher values of the target metric $a^i_t$ indicate better performance (e.g., CLIP score) or lower values are better (e.g., FID). The forgetting $F_t$ is defined as the maximum performance drop across tasks, calculated as:
\begin{equation}
F_t = 
\Bigg\{
\begin{aligned}
    &\max_{i \in \{0,...,T\}} \left( a^i_t - a^T_t \right), \text{if $a^i_t$ higher is better;}\\
    &\max_{i \in \{0,...,T\}} \left( a^T_t - a^i_t \right), \text{otherwise,}
\end{aligned}
\label{eq:fgt}
\end{equation}
where $a^i_t$ is the reward objective $R_t(\cdot)$ evaluated on the test dataset $C_{test}$ after training the diffusion model from task 0 to task $i$. Also, we define the original diffusion model before the fine-tuning process as the model in task 0. 

\section{Reward Incremental Distillation}
\label{sec:proposed}
Fig.~\ref{fig:teaser} illustrates the overall framework of our proposed~\ours. In this section, we introduce the two components of~\ours: the LoRA adapter group and momentum distillation. Then, we explain the optimization objective of~\ours.

\subsection{LoRA Adapter Group}
\label{subsec:baseline}
\begin{figure}[t]
  \centering
   \includegraphics[width=\linewidth]{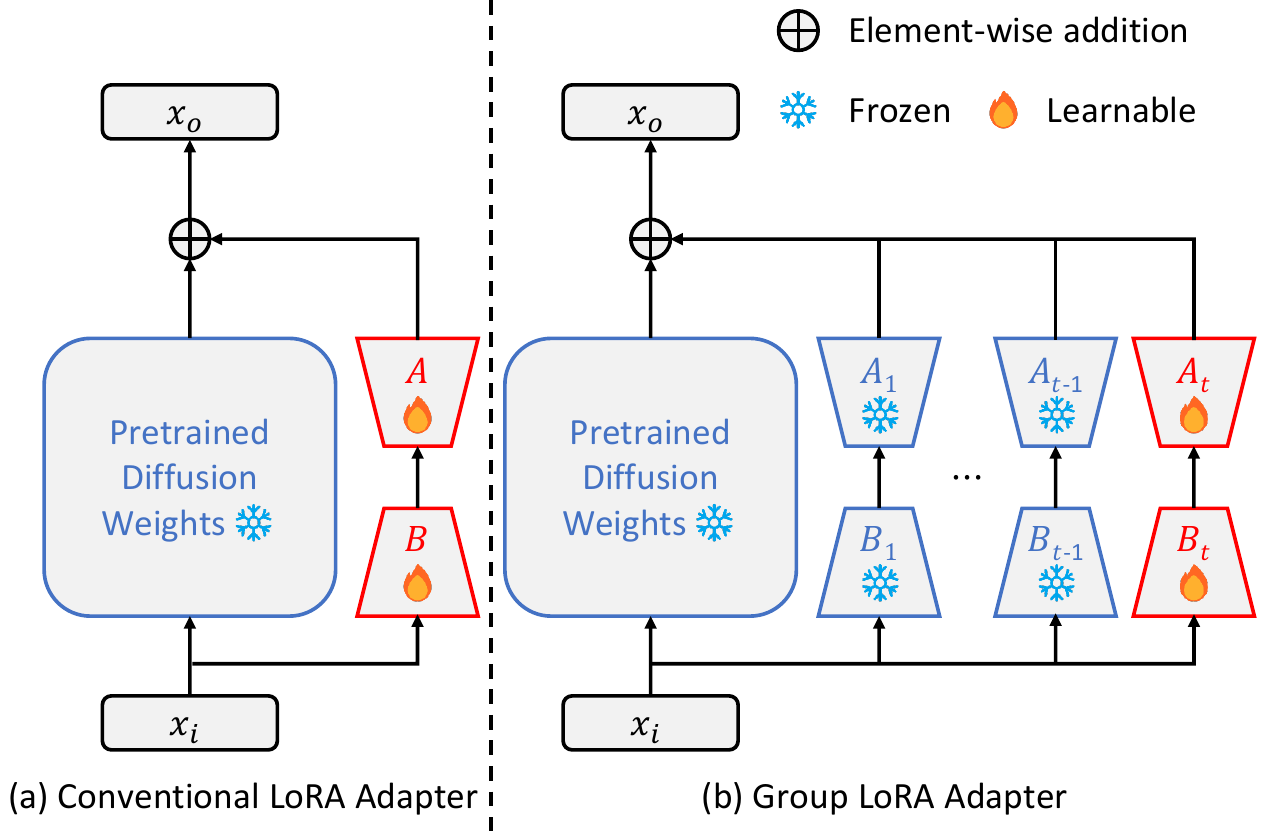}
   \vspace*{-10pt}
   \caption{Comparison of the conventional LoRA adapter and the proposed LoRA adapter group in a pretrained diffusion layer. Unlike the existing diffusion fine-tuning strategies that train a single pair of LoRA adapters, the adapter group expands and initializes a pair of new LoRA matrices when a new reward task arrives. The adapter group separates parameters for different reward tasks, for better knowledge retention.}
   \label{fig:lorainc}
   \vspace{-12pt}
\end{figure}

Instead of directly fine-tuning the model weights, inspired by InfLoRA~\cite{inflora}, we propose a structure of LoRA adapter group for the RIL setting as shown in Fig.~\ref{fig:lorainc}. When a new reward task $t \in \{1,...,T\}$ arrives, we expand each diffusion model layer $\mathcal{W}$ with a pair of newly initialized LoRA weight matrices $\mathcal{A}_t$ and $\mathcal{B}_t$. To avoid interference of learning task $t$ on previously acquired knowledge, we freeze the historical LoRA matrices $\mathcal{A}_i$ and $\mathcal{B}_i$ for $i < t$, along with the original diffusion weight $\mathcal{W}$. Thus, given the input $\bm{x_i}$ of the current layer, the feed-forward operation is calculated by:
\begin{equation}
\begin{aligned}
\bm{x_o} = \mathcal{W}\bm{x_i} + \sum_{i=1}^t \mathcal{A}_i\mathcal{B}_i\bm{x_i},
\end{aligned}
\label{eq:lorainc}
\end{equation}
where $\bm{x_o}$ is the output of the current layer. Following the standard LoRA initialization protocol, we use zero initialization for the dimension reduction matrix $B_t$ and Kaiming uniform initialization~\cite{kaiminginit} for the dimension expansion matrix $A_t$. 

\noindent\textbf{The advantage of LoRA adapter group.}
While training a single pair of LoRA adapters for all tasks is mathematically equivalent to training multiple task-specific adapters, a study~\cite{inflora} demonstrates that parameter sharing between new and previous tasks leads to interference, compromising the retention of previously learned knowledge. The proposed LoRA adapter group mitigates this issue by maintaining separate parameters for each task, thereby enhancing knowledge retention.

\noindent\textbf{Formulation of baseline.}
Since we are solving the RIL problem which has never been explored before, we derive a baseline method using the LoRA adapter group based on the existing state-of-the-art methods~\cite{alignprop, draft} for fine-tuning single reward objectives.

Following~\cite{alignprop, draft}, for computation efficiency, we use the DDIM~\cite{ddim} noise scheduler and truncate the backpropagation to the final sampling step. For the adapted baseline method on reward task $t$, we train the current LoRA adapter $\mathcal{A}_t,\mathcal{B}_t$ of the diffusion model with the following objective:
\begin{equation}
\begin{aligned}
\max_{\mathcal{A}_t,\mathcal{B}_t} \sum_{c \in C_{train}} R_t\left(f_{\mathcal{W},\mathcal{A},\mathcal{B}}\left(\bm{z}_N | c\right)\right),
\end{aligned}
\label{eq:baseobj}
\end{equation}
where $f_{\mathcal{W},\mathcal{A},\mathcal{B}}$ denotes the diffusion model parameterized by the original weights $\mathcal{W}$ and the LoRA weights $\mathcal{A},\mathcal{B}$.

\subsection{Momentum Distillation}
As shown in Fig.~\ref{fig:RILexample}, despite using the LoRA adapter group, the adapted baseline still suffers from the catastrophic forgetting problem. Inspired by the capability of knowledge distillation in alleviating forgetting in conventional incremental learning~\cite{icarl, mkd}, we propose to adapt the momentum distillation to the RIL setting in diffusion fine-tuning. To make this distillation feasible, we take advantage of the Exponential Moving Average (EMA) teacher~\cite{morales2024exponential} and adapt the distillation in the fine-tuning process. 

\noindent\textbf{The EMA teacher.}
The mean teacher~\cite{tarvainen2017mean} originally proposed the EMA-based teacher-student strategy, which has since been widely successful in self-supervised learning~\cite{moco, dino}. The advantage of applying the EMA teacher to the RIL problem is obvious: the EMA teacher model accumulates information from previous iterations, making it more resilient to the forgetting problem during the fine-tuning. Distillation between the EMA teacher and the student model (\textit{i.e.,} the diffusion model being fine-tuned) enhances robustness. Given a weight $\theta_{i}$ at $i$-th iteration in the student model $f_{\mathcal{W},\mathcal{A},\mathcal{B}}$, its counterpart $\theta_{i}^{T}$ in the EMA teacher model $f_{\mathcal{W},\mathcal{A},\mathcal{B}}^{T}$ is updated as:
\begin{equation}
\begin{aligned}
\theta_{i}^{T} =  \alpha \theta_{i-1}^{T} + \left(1 - \alpha\right) \theta_{i},
\end{aligned}
\label{eq:ema}
\end{equation}
where $\alpha$ is the momentum coefficient. 

\begin{figure}[t]
  \centering
   \includegraphics[width=.95\linewidth]{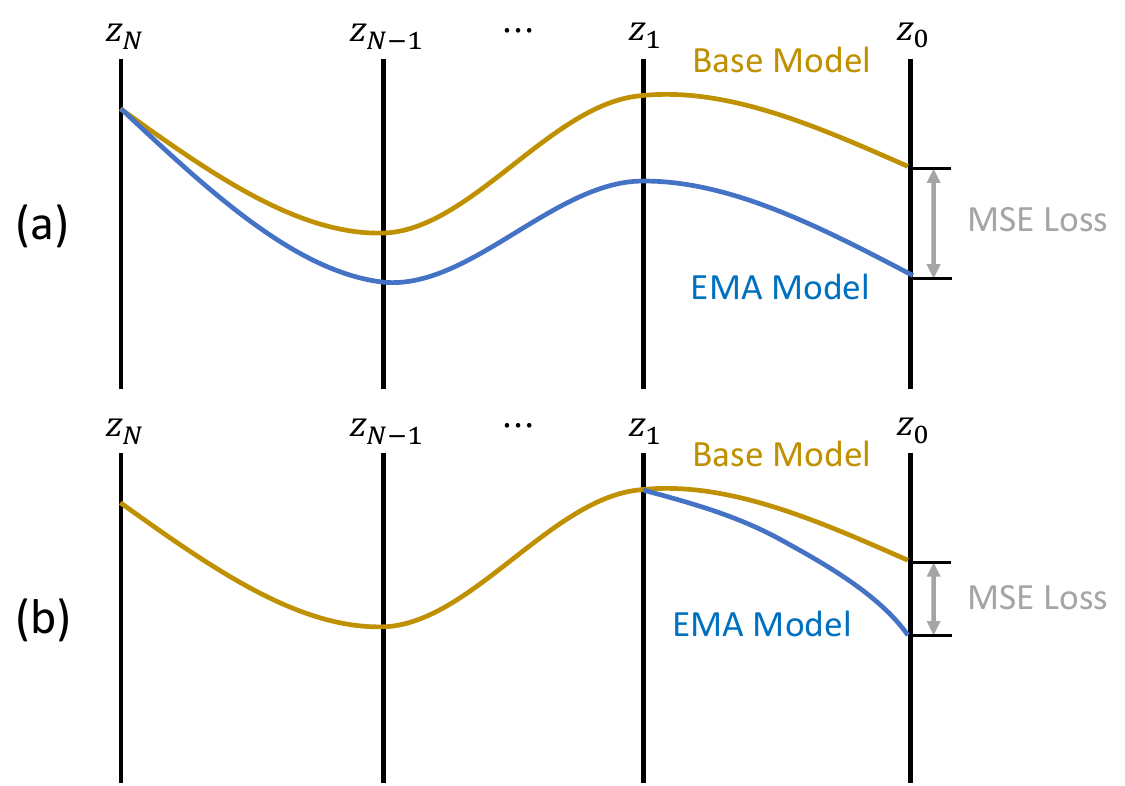}
   \vspace*{-6pt}
   \caption{Comparison between the naive full-step distillation strategy and the proposed last-step distillation strategy. (a) The full-step strategy aligns pixel-wise outputs across all denoising steps, starting from random noise $\bm{z}_N$, but suffers from high computational cost and error accumulation over time steps. (b) The last-step strategy only aligns the final diffusion step, reducing the computation and mitigating error accumulation, leading to a more efficient and stable fine-tuning process.}
   \label{fig:curve}
    \vspace{-9pt}
\end{figure}

\begin{figure*}
    \centering
    \includegraphics[width=\linewidth]{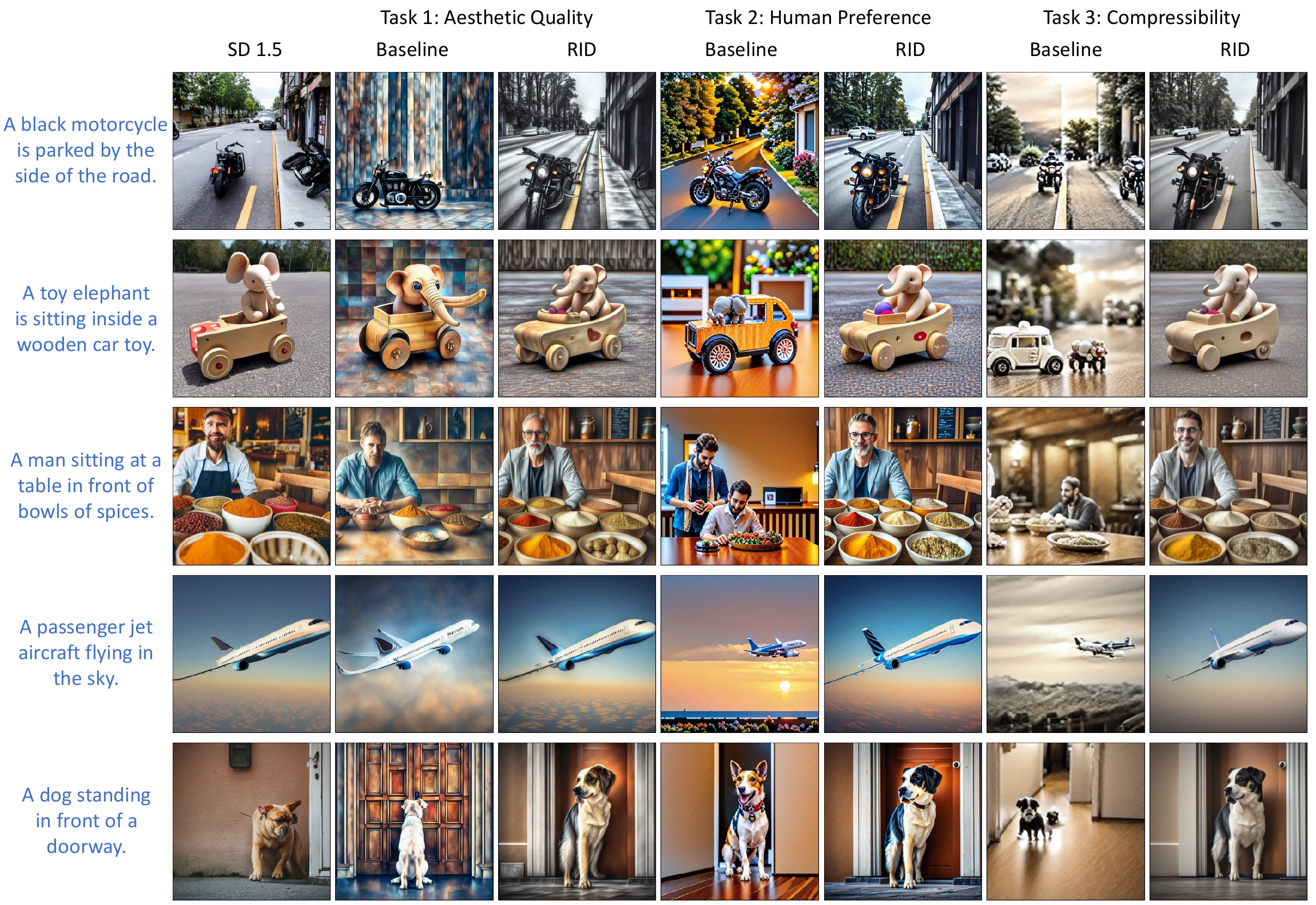}
    \vspace*{-14pt}
    \caption{Qualitative comparison of generation results using a fixed task sequence in the RIL setting. In generation, instead of reusing the prompts in the training dataset, we use novel prompts in the test dataset. From the figure, we can see that the adapted baseline suffers from a severe forgetting issue as more tuning tasks are introduced, while RID consistently improves upon the original Stable Diffusion model.}
    \label{fig:res}
    \vspace{-12pt}
\end{figure*}

\noindent\textbf{The EMA distillation.}
Using the EMA teacher $f_{\mathcal{W},\mathcal{A},\mathcal{B}}^{T}$, a straightforward approach to distillation is to minimize the pixel-wise $l_2$-norm between the image generated by both models, defined as: 
\begin{equation}
\begin{aligned}
\min_{\mathcal{A}_t,\mathcal{B}_t} \sum_{c \in C_{train}} \left|\left|f_{\mathcal{W},\mathcal{A},\mathcal{B}}(\bm{z}_N | c) - f_{\mathcal{W},\mathcal{A},\mathcal{B}}^T(\bm{z}_N | c)\right|\right|^2,
\end{aligned}
\label{eq:naive}
\end{equation}
where the diffusion model $f_{\mathcal{W},\mathcal{A},\mathcal{B}}(\bm{z}_N | c)$ denoise the random noise $\bm{z}_N \sim \mathcal{N}(0,\mathbf{I})$ to image output $\bm{z}_0$ conditioned by prompt $c \in C_{train}$. We refer to this as the full-step distillation strategy because both the EMA teacher and the student model start the denoising from the random noise $\bm{z}_N$. However, as shown in the diffusion trajectory in Fig.~\ref{fig:curve}, we find this strategy suboptimal. On one hand, full-step distillation doubles the computation cost of feed-forward propagation; on the other hand, the inconsistency between the student and teacher model accumulates over time steps $\{1,...,N\}$, and the gradient in the MSE loss can be interfered by this accumulated error. To address this issue, we propose last-step distillation, as shown in Fig.~\ref{fig:curve}. We use only the student model $f_{\mathcal{W},\mathcal{A},\mathcal{B}}$ for time steps $\{2,...,N\}$, obtaining the last-step latent $\bm{z}_1$. Then, we conduct the last step distillation as follows:
\begin{equation}
\begin{aligned}
\min_{\mathcal{A}_t,\mathcal{B}_t} \sum_{c \in C_{train}} ||f_{\mathcal{W},\mathcal{A},\mathcal{B}}(\bm{z}_1 | c) - f_{\mathcal{W},\mathcal{A},\mathcal{B}}^T(\bm{z}_1 | c)||^2,
\end{aligned}
\label{eq:laststep}
\end{equation}
where we truncate the backpropagation of $\bm{z}_1$ to the former time steps. Last-step distillation has two advantages. First, it only introduces a single extra forward propagation for the EMA teacher, amounting to just 2\% additional computation when the total denoising steps are set to 50. Also, last-step distillation avoids the accumulated error resulting from the inconsistencies in denoising by the EMA teacher and student at earlier time steps.

\begin{table*}
    \centering
    \resizebox{\textwidth}{!}{
    \begin{tabular}{cccccccccc}
\toprule
\multirow{2}{*}{Method} & \multicolumn{3}{c}{Tasks} & \multicolumn{4}{c}{Target Metrics (Forgetting metric)} & \multicolumn{2}{c}{General Metrics}  \\
\cmidrule(lr){2-4}
\cmidrule(lr){5-8}
\cmidrule(lr){9-10}
& Task 1 & Task 2 & Task 3 & Aesthetic score $\uparrow$ & HPS score $\uparrow$ & PNG Size (kB) $\downarrow$ & JPEG Size (kB) $\downarrow$ & CLIP score $\uparrow$ & Zero-shot FID $\downarrow$ \\
\midrule
Baseline & \multirow{2}{*}{Aesthetic} & \multirow{2}{*}{-} & \multirow{2}{*}{-}
        &	6.07 {(-)}&	   0.257 {(-)}&	409.78 {(-)}&	55.73 {(-)}&	22.74 {(3.03)}&	85.53 {(13.99)}\\
\ours&&&&	6.03 {(-)}&	   0.267 {(-)}&	443.57 {(-)}&	60.41 {(-)}&	25.76 \textbf{(0.01)}&	79.40 \textbf{(7.86)}\\
\midrule
Baseline & \multirow{2}{*}{Aesthetic} & \multirow{2}{*}{HPS} & \multirow{2}{*}{-}
        &	5.79 {(0.28)}&	0.282 {(-)}&	415.96 {(-)}&	58.91 {(-)}&	23.19 {(2.58)}&	89.71 {(18.17)}\\
\ours&&&&	5.94 \textbf{(0.09)}& 0.276 {(-)}&		451.70 {(-)}&	68.01 {(-)}&	25.46 \textbf{(0.31)}&	80.04 \textbf{(8.50)}\\
\midrule
Baseline & \multirow{2}{*}{Aesthetic} & \multirow{2}{*}{HPS} & \multirow{2}{*}{Compress}
        &	4.98 {(1.09)}&	0.249 {(0.033)}&	316.89 {(-)}&	34.20 {(-)}&	22.39 {(3.38)}&	104.85 {(33.31)}\\
\ours&&&&	5.55 \textbf{(0.48)}&	0.272 \textbf{(0.004)}&	373.63 {(-)}&	40.50 {(-)}&	25.00 \textbf{(0.77)}&	80.28 \textbf{(8.74)}\\

\bottomrule
\end{tabular}
    }
    \vspace*{-8pt}
    \caption{Performance of the diffusion model after sequential fine-tuning on a fixed series of tasks: aesthetic quality, human preference (HPS), and compressibility. We evaluate the performance of the model at the conclusion of each task. The numbers in the parentheses are the forgetting metric (lower is better), as defined in Eq.~\ref{eq:fgt}.
    }
    \label{tab:incre_res}
    \vspace{-8pt}
\end{table*}

\begin{table*}
    \centering
    \resizebox{\textwidth}{!}{
    \begin{tabular}{cccccccccc}
\toprule
\multirow{2}{*}{Method} & \multicolumn{3}{c}{Tasks} & \multicolumn{4}{c}{Target Metrics  (Forgetting metric)} & \multicolumn{2}{c}{General Metrics}  \\
\cmidrule(lr){2-4}
\cmidrule(lr){5-8}
\cmidrule(lr){9-10}
& Task 1 & Task 2 & Task 3 & Aesthetic score $\uparrow$ & HPS score $\uparrow$ & PNG Size (kB) $\downarrow$ & JPEG Size (kB) $\downarrow$ & CLIP score $\uparrow$ & Zero-shot FID $\downarrow$ \\
\midrule
Baseline & \multirow{2}{*}{Aesthetic} & \multirow{2}{*}{HPS} & \multirow{2}{*}{Compress}
        &	4.98 {(1.09)}&	0.249 {(0.033)}&	316.89 {(-)}&	34.20 {(-)}&	22.39 {(3.38)}&	104.85 {(33.31)}\\
\ours&&&&	5.55 \textbf{(0.48)}&	0.272 \textbf{(0.004)}&	373.63 {(-)}&	40.50 {(-)}&	25.00 \textbf{(0.77)}&	80.28 \textbf{(8.74)}\\
\midrule
Baseline & \multirow{2}{*}{Aesthetic} & \multirow{2}{*}{Compress} & \multirow{2}{*}{HPS}
&	5.28 {(0.79)}&	0.280 {(-)}&	422.34 {(36.71)}&	69.56 {(24.53)}&	22.26 {(3.51)}&	119.30 {(47.76)}\\
\ours&&&&	5.40 \textbf{(0.63)}&	0.281 {(-)}&	369.05 \textbf{(15.64)}&	44.09 \textbf{(7.21)}&	24.36 \textbf{(1.41)}&	82.89 \textbf{(11.35)}\\
\midrule
Baseline & \multirow{2}{*}{Compress} & \multirow{2}{*}{HPS} & \multirow{2}{*}{Aesthetic}
&	7.23 {(-)}&	0.265 {(0.026)}&	307.13 {(0.00)}&	37.58 \textbf{(0.00)}&	18.92 {(6.85)}&	124.57 {(53.03)}\\
\ours&&&&	5.77 {(-)}&	0.275 \textbf{(0.014)}&	343.71 {(0.00)}&	40.83 {(0.08)}&	24.43 \textbf{(1.50)}&	91.04 \textbf{(19.50)}\\
\midrule
Baseline & \multirow{2}{*}{Compress} & \multirow{2}{*}{Aesthetic} & \multirow{2}{*}{HPS}
&	5.25 {(0.67)}&	0.279 {(-)}&	412.77 {(35.78)}&	62.99 {(22.88)}&	22.02 {(3.75)}&	116.20 {(44.66)}\\
\ours&&&&	5.19 \textbf{(0.04)}&	0.283 {(-)}&	389.83 \textbf{(3.46)}&	47.15 \textbf{(6.40)}&	24.72 \textbf{(1.21)}&	76.99 \textbf{(5.45)}\\
\midrule
Baseline & \multirow{2}{*}{HPS} & \multirow{2}{*}{Aesthetic} & \multirow{2}{*}{Compress}
&	4.76 {(1.81)}&	0.227 {(0.058)}&	333.32 {(-)}&	29.35 {(-)}&	16.58 {(9.19)}&	134.98 {(63.44)}\\
\ours&&&&	5.66 \textbf{(0.23)}&	0.274 \textbf{(0.011)}&	327.52 {(-)}&	37.59 {(-)}&	24.06 \textbf{(1.71)}&	92.28 \textbf{(20.74)}\\
\midrule
Baseline & \multirow{2}{*}{HPS} & \multirow{2}{*}{Compress} & \multirow{2}{*}{Aesthetic}
&	6.18 {(-)}&	0.251 {(0.034)}&	429.12 {(76.56)}&	51.50 {(11.46)}&	16.41 {(9.36)}&	185.61 {(114.07)}\\
\ours&&&&	5.75 {(-)}&	0.275 \textbf{(0.010)}&	326.05 \textbf{(8.55)}&	38.55 \textbf{(0.57)}&	23.93 \textbf{(1.84)}&	92.19 \textbf{(20.65)}\\
\midrule
\multicolumn{4}{c}{Average (Baseline)} & 5.62 & 0.258  & 370.26 & 47.53 & 19.76 & 130.92 \\
\multicolumn{4}{c}{Average (\ours)} & 5.55 & 0.277  & 354.96 & 41.45 & 24.42 & 85.95 \\
\multicolumn{4}{c}{Stable Diffusion V1.5} & 5.23 & 0.259  & 425.13 & 50.40 & 25.77 & 71.54 \\
\multicolumn{4}{c}{Model Soup} & 5.56 & 0.275  & 399.15 & 47.36 & 25.23 & 86.24 \\
\bottomrule
\end{tabular}

    }
    \vspace*{-8pt}
    \caption{Final performance of incrementally fine-tuning the Stable Diffusion v1.5 model on different reward sequences. The numbers in the parentheses are the forgetting metric (lower is better), as calculated by Eq.~\ref{eq:fgt}. We also included the performance by mixing the weights with model soup~\cite{soup}.}
    \label{tab:main}
    \vspace{-14pt}
\end{table*}

\noindent\textbf{Optimization objective.}
The overall optimization objective of~\ours~is the combination of reward optimization as in Eq.~\ref{eq:baseobj} and EMA distillation as in Eq.~\ref{eq:laststep}. To improve computational efficiency, we also truncate the backpropagation of reward optimization to the last time step. Therefore, the optimization objective of~\ours~can be expressed as:
\begin{equation}
\begin{aligned}
\max_{\mathcal{A}_t,\mathcal{B}_t} \sum_{c \in C_{train}} (&R_t(f_{\mathcal{W},\mathcal{A},\mathcal{B}}(\bm{z}_1 | c)) - \\
& \lambda ||f_{\mathcal{W},\mathcal{A},\mathcal{B}}(\bm{z}_1 | c) - f_{\mathcal{W},\mathcal{A},\mathcal{B}}^T(\bm{z}_1 | c)||^2),
\end{aligned}
\label{eq:oursobj}
\end{equation}
where $\lambda$ is the balancing hyperparameter, which is set to 0.1 for all experiments based on our hyperparameter search. 

\section{Experiments}
\label{sec:experiments}

\subsection{Experiment Setup.}
\label{subsec:setup}

\noindent\textbf{The prompt dataset.} In our work, we fine-tune the Stable Diffusion V1.5~\cite{ldm} model with the real user prompts collected in the Human Preference Dataset (HPD)~\cite{hps}. Following~\cite{alignprop}, to validate the capability of~\ours~in generalizing to unseen prompts, we use 750 prompts in the HPD dataset for training and 50 prompts for evaluation. Additional detailed information and examples of this dataset are provided in the appendix.

\noindent\textbf{Evaluation metrics.}
After training the diffusion model on a given target sequence in the RIL problem, we use prompts $c$ from the test prompt dataset $C_{test}$ to generate test images. The test images are evaluated with target reward objectives, together with the general image quality metrics. The target reward objectives include: 
\begin{itemize}
    \item \textbf{Aesthetic quality.} The prediction of LAION aesthetic predictor v2~\cite{laionaesthetics2022}, averaged on all test images.

    \item \textbf{Human preference score.} The average prediction of the HPSv2~\cite{hps} predictor. 

    \item \textbf{Compressibility.} We evaluate the compressibility using two metrics: average file size of lossless compression (in PNG format) and average file size of lossy compression (in JPEG format at quality = 80). 
\end{itemize}
Moreover, we use two well-known general metrics to evaluate the overall quality of the generated image, including:
\begin{itemize}
    \item \textbf{Text-to-image alignment.} We measure the text-to-image alignment of the generated images as an overall quality metric using the CLIP score~\cite{clipscore}.

    \item \textbf{General natureness.} General visual naturalness of generated images is assessed using the Fréchet Inception Distance (FID)~\cite{heusel2017gans} on the MS-COCO dataset in a zero-shot setting.

\end{itemize}
Besides the evaluated metrics, we calculate the forgetting of each metric according to Eq.~\ref{eq:fgt} to measure the robustness of the methods in the fine-tuning process. 

\noindent\textbf{Implementation details.}
Following~\cite{alignprop}, we use the Stable Diffusion V1.5 as the base model for both methods. The rank in the group LoRA adapter is set to 4. Depending on the reward task, we assign different hyperparameter configurations after the hyperparameter search as detailed in the appendix. In the training, we conduct the experiments with two A100 GPUs and reported the training time in Sec.~\ref{sec:discussion}. For evaluation, we generate 200 test images with varying initial noises for each prompt in a test dataset of 50 prompts. To ensure a fair comparison, all test images are generated with the same random seed, maintaining consistency in both prompts and initial noise $\bm{z}_N$ across evaluations.

\subsection{Experimental Results}
\label{subsec:incre}
To investigate the model’s performance and stability across each stage, as well as the impact of cumulative tuning on image quality and adaptability, we first present results for one example task sequence. Then, we show the quantitative results of all task-incremental settings.

Table~\ref{tab:incre_res} presents the metric-wise results for fine-tuning the diffusion model on the example task sequence: aesthetic tuning first, followed by human preference (HPS) tuning, and finally compressibility tuning. The table shows that both the adapted baseline and RID achieve satisfactory performance on the target objective after each task. However, the adapted baseline exhibits more severe forgetting compared to RID. For instance, after fine-tuning on the HPS task, the adapted baseline’s aesthetic score drops from 6.07 to 5.79, resulting in a forgetting value (cf. Eq.~\ref{eq:fgt}, lower is better) of 0.28, while RID’s aesthetic score decreases from 6.03 to 5.94, with a forgetting value of 0.09. This trend is consistent across general metrics, including the CLIP score and zero-shot MS-COCO FID. As shown in the table, the forgetting effect becomes more pronounced as more tasks are introduced, but RID consistently outperforms the adapted baseline with less forgetting. More interestingly, by the end of fine-tuning, the aesthetic and HPS quality of the adapted baseline falls below that of the original Stable Diffusion, while RID maintains significant improvements across all target metrics.


Fig.~\ref{fig:res} provides qualitative results, showcasing sample generations throughout the reward incremental learning process. Consistent with the metric-wise results, both methods show notable improvement in the aesthetic quality of generated samples following the first task (aesthetic tuning). However, as more tasks are added, the quality of the adapted baseline declines significantly, with a marked occurrence of visual structure-wise forgetting, while RID still maintains a high fidelity in generation. 

In Table~\ref{tab:main}, We also report the metric-wise performance at the end of fine-tuning for different task orders, alongside the performance of the original Stable Diffusion V1.5 and the results obtained using model soup~\cite{soup} (cf. Sec.~\ref{sec:discussion}). In all settings, the adapted baseline consistently shows significant forgetting, while RID achieves notable improvements across all target metrics with alleviated forgetting. Additional qualitative results are provided in the appendix.

\section{Discussion}
\label{sec:discussion}

\begin{table}
    \centering
    \resizebox{.7\linewidth}{!}{
    \begin{tabular}{cccc}
\toprule
\multirow{2}{*}{Method} & \multicolumn{3}{c}{Training time (min / epoch)} \\
\cmidrule(lr){2-4}
 & Aesthetic & HPS & Compress \\
\midrule
Baseline & 10.05 & 9.93 & 10.04\\
\ours    & 10.10 & 10.02 & 10.17\\
\bottomrule
\end{tabular}

    }
    \caption{Comparison of the average training time per epoch between~\ours~and the adapted baseline method across various reward tasks. Compared with the training time of the baseline, the additional computation overhead introduced by RID is minor.}
    \label{tab:time}
    \vspace{-10pt}
\end{table}

\noindent\textbf{Computational overhead.}
Compared with our adapted baseline,~\ours~introduces two extra sources of forward computation: one extra denoising step because of last-step distillation, as mentioned in Sec.~\ref{sec:proposed}, and one extra variational autoencoder decoding step. However, we argue that these computations contribute minimally to the overall fine-tuning overhead. To validate the computational efficiency of our proposed method, we compare the training time of~\ours~with that of the baseline method described in Sec.~\ref{sec:preliminary}. We calculate the average time per epoch for both methods when trained on the various reward tasks on two Nvidia A100 GPUs. As shown in Table~\ref{tab:time}, the extra computational overhead introduced by~\ours~is less than 1\% and thus negligible.

\noindent\textbf{Comparison with model soup.}
Before the RIL setting, a common approach for multi-reward optimization was to combine the weights of multiple models, each independently fine-tuned on a specific reward, in a technique known as model soup~\cite{soup}. This approach requires training separate models for each reward, leading to increased training complexity and memory overhead. In contrast, the RIL setting seeks to optimize a single model incrementally across multiple rewards, maintaining adaptability without the need to train and store multiple models. To perform the model soup mixing, we average the weights from three models $\theta_1, \theta_2, \theta_3$ fine-tuned with the adapted baseline on different reward functions by: $\theta_{soup} = \alpha \cdot \theta_1 + \beta \cdot \theta_2 + (1-\alpha-\beta) \cdot \theta_3$, where $\alpha, \beta$ are two scaler mixing coefficients. We present a comparison of the generation results using model soup with $\alpha = \beta = 0.333$ in Table~\ref{tab:main}. While model soup yields improvements across all target metrics compared to the Stable Diffusion baseline, it shows limited performance on specific attributes, such as compressibility. We attribute this to a knowledge gap, as each model in the soup is tuned independently, without the knowledge of the others. In contrast, RID directly tunes the group LoRA adapter on the reward objectives, using \textit{optimization} rather than simple \textit{mixing} to adjust model parameters. As a cohesive approach, RID not only preserves memory efficiency by maintaining a single model but also yields more robust performance across multiple reward tasks. 

\section{Conclusion}
\label{sec:conclusion}

In this paper, we introduce the Reward Incremental Learning (RIL) problem, which focuses on fine-tuning a single diffusion model across a sequentially expanding set of reward tasks. Moreover, we adapt the state-of-the-art fine-tuning methods to the RIL problem, observing a salient forgetting issue as additional reward tasks are introduced. To address this, we propose the Reward Incremental Distillation (RID) and achieve a better adaptability-robustness tradeoff. Both qualitative and quantitative experimental results demonstrate the effectiveness of RID. In future work, we plan to expand the evaluation to include a broader range of reward tasks.
{
    \small
    \bibliographystyle{ieeenat_fullname}
    \bibliography{main}
}

\clearpage
\setcounter{page}{1}
\maketitlesupplementary


\begin{figure*}[t]



       
    \includegraphics[width=\linewidth]{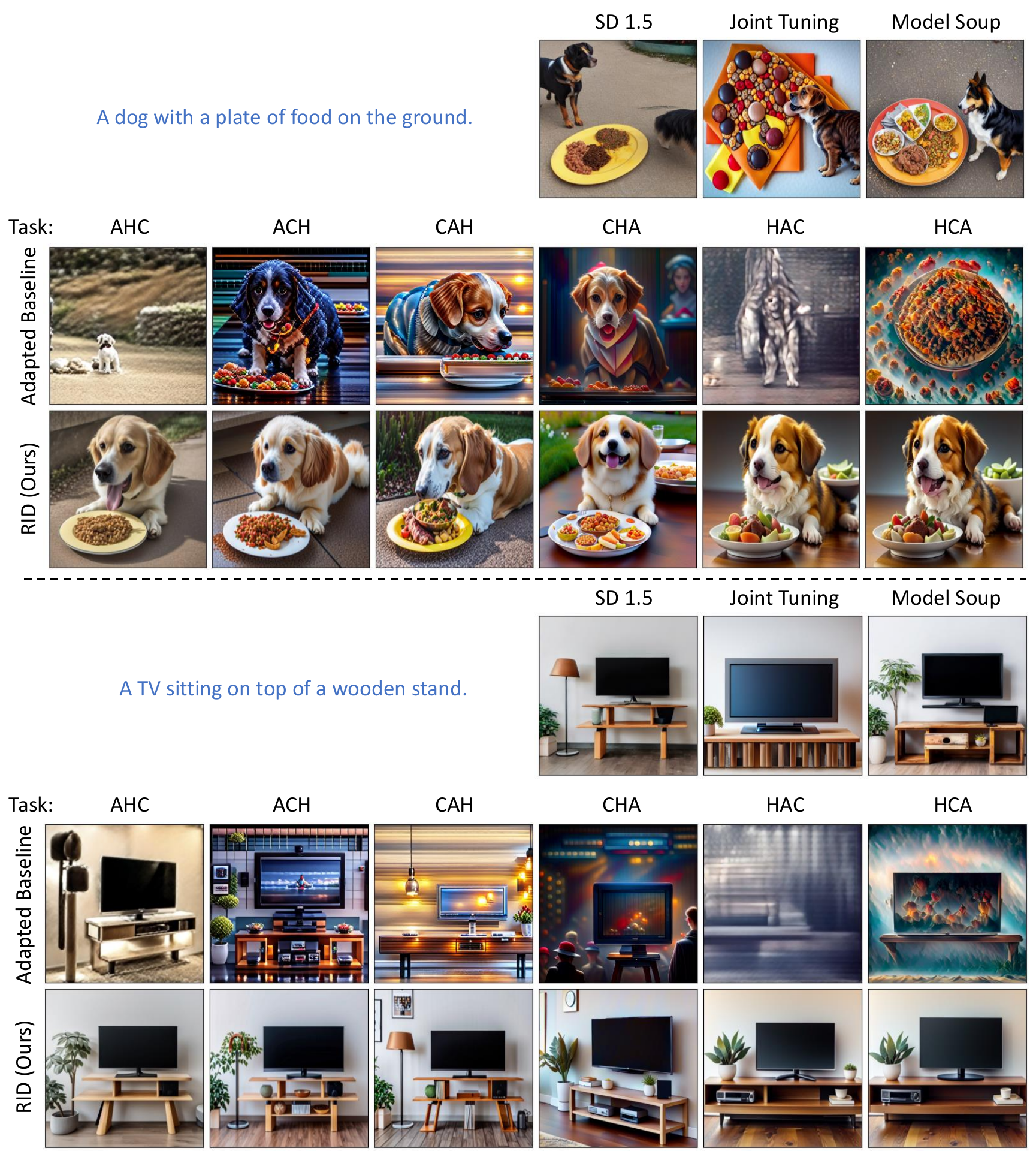}
    \caption{Qualitative comparison of generation results at the end of tuning on different task sequences, along with the results of joint tuning and model soup. For simplicity, we use abbreviations of the target objective to express task orders (\textit{e.g.,} ``AHC'' represents aesthetic tuning first, followed by human preference tuning, and finally compressibility tuning.). }
    \vspace{20pt}
    \label{fig:extra}
\end{figure*}

\begin{spacing}{0.8}
\begin{algorithm}[t]
\scriptsize
\begin{minted}{python}
# model: diffusion model with group LoRA Adapter
# ema_model: EMA teacher model
# train_set: prompt dataset
# reward: reward function

While not converged:
    prompt = sample(train_set)
    z = initial_noise_sample()
    for n in range(N, 0, -1): # DDIM scheduler
        if n > 1:
            with torch.no_grad():
                z = model(z, n, prompt)
        else:
            with torch.no_grad():
                z_tea = ema_model(z, n, prompt)
            z_0 = model(z, n, prompt)

    loss = - reward(z_0) + \
             lam * mse_loss(z_0, z_tea) # Eq. 8
    loss.backward()
    optimizer.step()
    ema_model.update(model)
        
\end{minted}
\vspace{-4pt}
\caption{PyTorch-like pseudo-code of the EMA distillation in our proposed~\ours.}
\label{code:pseudo_code}
\end{algorithm}
\end{spacing}

\section{Pseudo Code for EMA Distillation}
For better understanding, we present the pseudo-code for the EMA distillation as in Alg.~\ref{code:pseudo_code}. For simplicity, we ignored the autoencoder decoding step at the end of the diffusion.

\section{Further Analysis and Qualitative Samples}

\noindent\textbf{Comparison with joint tuning.} 
Apart from the RIL setting, we also compare results with a jointly trained approach that optimizes the weighted sum of three reward objectives using the adapted baseline. In this setup, the joint reward $R_j$ is computed by:
\begin{equation}
\begin{aligned}
R_j = 0.01 \times R_a + 2 \times R_h + R_c,
\end{aligned}
\label{eq:joint}
\end{equation}
where $R_a$, $R_h$, and $R_c$ are the aesthetic reward, human preference reward, and compressibility reward, respectively. We train the joint reward for 20 epochs, and the final quantitative results are shown in Table~\ref{tab:extra}. We chose such weighting because different rewards have different gradient norms, and we conducted a small-scale hyperparameter search based on the generation quality with different weight coefficients. 
As shown in Table~\ref{tab:extra}, when trained with the joint loss, all of the target metrics show improvement compared with the Stable Diffusion V1.5 baseline. Nevertheless, the improvement in the target objective is not so pronounced compared with the average improvement of RID. Moreover, the qualitative results are shown in Fig.~\ref{fig:extra}. 

Our experiments revealed two key limitations of the joint tuning strategy. First, due to the varying scales of gradients across reward tasks, balancing the weight coefficients in Eq.~\ref{eq:joint} requires careful tuning, adding complexity to the hyperparameter search. Second, the optimization process is often hindered by conflicting target rewards. For instance, the aesthetic reward and compressibility reward frequently conflict, resulting in suboptimal performance on target objectives compared to the RIL tuning approach. 


\noindent\textbf{Extra qualitative results.}
As mentioned in Sec.~\ref{subsec:incre}, we present extra qualitative results at the end of the tuning for different task orders, along with the results of joint tuning and model soup, as shown in Fig.~\ref{fig:extra}. Similar to our observation in Sec.~\ref{subsec:incre}, RID can generate high-fidelity images that are well aligned with the target objective, while the adapted baseline has limited generation quality. Moreover, compared to joint tuning and model soup, RID not only achieves superior objective quality (\textit{i.e.,} aesthetic quality, human preference, compressibility) but also produces more natural and coherent generations.

\begin{table}
    \centering
    \resizebox{\linewidth}{!}{
    \begin{tabular}{l}
\toprule
\multicolumn{1}{c}{Prompt Examples} \\
\midrule
Fruit in a jar filled with liquid sitting on a wooden table. \\
A passenger jet aircraft flying in the sky. \\
Several people standing next to each other that are snow skiing. \\ 
A passenger jet being serviced on a runway in an airport. \\
Three people are preparing a meal in a small kitchen. \\
A pair of planes parked in a small rural airfield. \\
A bathroom with a stand alone shower and a peep window. \\
Several vehicles with pieces of luggage on them with planes off to the side.\\
A black motorcycle is parked by the side of the road.\\
A small bathroom with a tub, toilet, sink, and a laundry basket are shown.\\
A bus stopped on the side of the road while people board it.\\
A bunch of people posing with some bikes.\\
\bottomrule
\end{tabular}

    }
    \caption{Examples of the prompts in the HPD dataset.}
    \label{tab:prompts}
\end{table}

\begin{table*}
    \centering
    \resizebox{.85\textwidth}{!}{
    \begin{tabular}{ccccccc}
\toprule
\multirow{2}{*}{Method} & \multicolumn{4}{c}{Target Metrics} & \multicolumn{2}{c}{General Metrics}  \\
\cmidrule(lr){2-5}
\cmidrule(lr){6-7}
& Aesthetic score $\uparrow$ & HPS score $\uparrow$ & PNG Size (kB) $\downarrow$ & JPEG Size (kB) $\downarrow$ & CLIP score $\uparrow$ & Zero-shot FID $\downarrow$ \\
\midrule
\multicolumn{1}{c}{Stable Diffusion V1.5} & 5.23 & 0.259  & 425.13 & 50.40 & 25.77 & 71.54 \\
\multicolumn{1}{c}{Joint Tuning} & 5.43 & 0.263 & 378.99 & 49.20 & 20.00 & 82.52 \\
\multicolumn{1}{c}{Average (\ours)} & 5.55 & 0.277  & 354.96 & 41.45 & 24.42 & 85.95 \\
\bottomrule
\end{tabular}

    }
    \caption{The comparison of the final performance of joint tuning and RID.}
    \label{tab:extra}
\end{table*}

\section{Experiment Details}

\noindent\textbf{Hyperparameter details.}
We use different hyperparameters, specifically epochs and learning rates, for fine-tuning each reward task due to variations in the gradients generated by the reward functions. Specifically, for all experiments, we train for 120 epochs for human preference tuning, 15 epochs for compressibility, and 10 epochs for aesthetic tuning. For the adapted baseline, following~\cite{alignprop, draft}, we employ a large batch size of 128 and a higher learning rate of $10^{-3}$ across all experiments. In contrast, RID uses an EMA teacher, which benefits from longer optimization. Consequently, we adopt a smaller batch size of 8 and a reduced learning rate of $5 \times 10^{-5}$. Under this configuration, the adapted baseline performs worse. 

\noindent\textbf{Details about the prompt dataset.}
As mentioned in Sec.~\ref{subsec:setup}, we use the prompts from the Human Preference Dataset (HPD)~\cite{hps}. HPD comprises four styles of prompts: including ``animation'', ``painting'', ``concept-art'', and ``photo'', with each style containing 800 prompts. In our experiments, follow~\cite{alignprop}, we use the prompts from the ``photo'' domain, using its prompts for both training and evaluation. For the train-test split, we follow the implementation of~\cite{alignprop}, using the allocated 750 prompts for training and 50 prompts for testing. Some examples of the prompts are shown in Table~\ref{tab:prompts}.

\end{document}